\documentclass[conference]{IEEEtran}
\usepackage[latin1]{inputenc}
\usepackage{subfigure}
\usepackage{amsmath}
\usepackage{amssymb}
\usepackage{cite}
\usepackage{multirow}
\ifCLASSINFOpdf
\usepackage[pdftex]{graphicx}
\graphicspath{{./pdf/}{./jpeg/}}
   \DeclareGraphicsExtensions{.pdf,.jpeg,.png}
\else
  \usepackage[dvips]{graphicx}
  \graphicspath{{../eps/}}
  \DeclareGraphicsExtensions{.eps}
\fi

\usepackage{url}

\begin{document}

\title{Cloud-based Evolutionary Algorithms: An algorithmic study}
\author{\IEEEauthorblockN{ Juan-J. Merelo, Maribel García-Arenas,\\  
Antonio M. Mora,  Pedro Castillo,
\\ Gustavo Romero, JLJ Laredo}
\IEEEauthorblockA{Departamento de Arquitectura y Tecnolog\'ia de Computadores \\
University of Granada\\
Email: \{maribel,jmerelo,amorag,pedro,gustavo,juanlu\}@geneura.ugr.es}
}

\maketitle
\begin{abstract}
After a proof of concept using Dropbox\texttrademark, a free
storage and synchronization, showed that an evolutionary algorithm
using several dissimilar computers
connected via  WiFi or Ethernet had a good scaling behavior in terms
of evaluations per second, it remains to be proved whether that effect
also translates to the algorithmic performance of the algorithm. In
this paper we will check several different, and difficult, problems,
and see what effects the automatic load-balancing and asynchrony have
on the speed of resolution of problems.

\end{abstract}

\IEEEpeerreviewmaketitle

\section{Introduction}

The main objective of this research is to find easily available means
to either use or connect computing nodes in a distributed evolutionary
computation experiment, and this often means resorting to free and
readily available services. Dropbox\texttrademark is one of these
services: it is commercialized as a {\em cloud storage} service, which
is free up to a certain level of use (measured in traffic and
usage). There are many other services like this one; however Dropbox
was chosen due to its popularity, which also implies having many more
potential volunteer users of a massive evolutionary computation
experiment. there are also other features that make it the right tool
for these experiments. Some other cloud storage services, like Wuala,
provide a client program on which one must add explicitly the files
that will be stored, which does not allow a seamless integration with
the filesystem; others, like ZumoDrive, use remotely-mounted
filesystems whose access is not so fast. Dropbox monitors local
filesystems, and uploads them asynchronously, which makes it faster
from the local point of view\footnote{The characteristics of these and
  others online backup services can be seen in \url{http://en.wikipedia.org/wiki/Comparison_of_online_backup_services}}. 

In the experiments we are performing, we are interested in its use as
a file {\em synchronization} service. When one file in one of the
folders that is monitored by Dropbox is changed, it is uploaded to
Dropbox servers and then distributed to all the clients that share the
same folder. It is interesting, however, to note that from the
programming point of view, all folders are written and read as local
one, which makes its use quite easy, and also seamless. 

In previous experiments \cite{dropbox:cec} we measured whether adding several computers
to an experiment of this kind resulted in an increase in the number of
simultaneous evaluations. In this paper we will measure whether
besides an increase in speed, the algorithm profits from the
distribution and asynchrony of the particular instance we have
implemented, or on the contrary it suffers from it. In order to do
that, we chose two optimization problems with a different degree of difficulty, and measured the time needed to find the solution, along with the number of evaluations. 

The rest of the paper is organized as follows: after a brief section presenting the state of the art in voluntary and pool-based evolutionary computation,  we describe the 
algorithm, the experimental setup and the implementation in Section
\ref{sec:desc}; results of these experiments will be briefly presented
in Section \ref{sec:res}, to be followed by the conclusion, discussion and future lines of
work in Section \ref{sec:conc}. 

\section{State of the Art}

Cloud computing \cite{Armbrust:2010:VCC:1721654.1721672,springerlink:10.1007/978-1-4419-6524-0_14} is an
emergent technology, and as such research related to it is just
recently emerging. Research addressing cloud storage is mainly related
to content delivery \cite{Broberg20091012} or designing data
redundancy schemes to ensure information integrity
\cite{PamiesJuarez2010}. However, its use in distributed computing has
not been addressed in such depth. Even if it is related to data grids
\cite{chervenak2000data}, in this paper we address the use of free cloud
storage as a medium for doing distributed evolutionary computation, in
a more or less parasitic way \cite{Barabasi2001Parasitic}, since we
use the infrastructure laid by the provider as part of an immigration
scheme in an island-based evolutionary algorithm \cite{GA_Island_Model}.

Thus we will have to look at pool-based distributed evolutionary
algorithms for the closest methods to the one presented here. In these
methods, several nodes or {\em islands} share a {\em pool} where the common 
information is written and read. To work against a single pool of solutions is an idea that has been considered almost from the beginning of research in distributed evolutionary algorithms. Asynchronous Teams or A-Teams
\cite{de1991genetic,talukdar1998asynchronous,talukdar2003asynchronous}
were proposed in the early nineties as a cooperative scheme for
autonomous agents. The basic idea is to create a work-flow on a set of
solutions and apply several heuristic techniques to improve
them, possibly including humans working on them. This technique is not
constrained to evolutionary algorithms, since it can be applied to any
population based technique, but in the context of EAs, it would mean creating
different single-generation algorithms, with possibly several
techniques, that would create a new generation from the existing
pool. 

The A-Team method does not rely on a single implementation, focusing
on the algorithmic and data-flow aspects, in the same way as the
Meandre \cite{llora51meandre} system, which creates a data flow
framework, with its own language (called ZigZag), which can be
applied, in particular, to evolutionary algorithms.

While algorithm design is extremely important, implementation issues
always matter, and some recent papers 
have concentrated on dealing with pool architectures in a single
environment: G. Roy et al. \cite{pool:ga} propose a shared memory
multi-threaded architecture, in which several threads work
independently on a single shared memory, having read access to the
whole pool, but write access to just a part of it. That way, interlock
problems can be avoided, and, taking advantage of the multiple
thread-optimized architecture of today's processors, they can obtain
very efficient, running time-wise, solutions, with the added algorithmic
advantage of working on a distributed environment. Although they do
not publish scaling results, they discuss the trade off of working with
a pool whose size will have a bigger effect on performance than the
population size on single-processor or distributed EAs. The same
issues are considered by Bollini and Piastra in
\cite{bollini1999distributed}, who present a design pattern for
persistent and distributed evolutionary algorithms; although their
emphasis is on persistence, and not performance, they try to present
several alternatives to decouple population storage from evolution
itself ({\em traditional} evolutionary algorithms are applied 
directly on storage) and achieve that kind of persistence, for which they
propose an object-oriented database management system accessed from a
Java client. In this sense, our former take on browser-based evolutionary
computation \cite{agajaj} 
% Antonio - quitar los autores de AGAJAJ
is also similar, using for persistence a small database accessed through a web interface, but only for the purpose of interchanging individuals among the different nodes, not as storage for the whole population. 

In fact, the efforts mentioned above have not had much continuity, probably
due to the fact that there have been, until now, few (if any) publicly
accessible online databases. However, given the rise of cloud
computing platforms over the last few years, interest in this kind of
algorithms has bounced back, with implementations using the public
FluidDB platform \cite{merelo2010fluid} having been recently
published.

\section{Description of the algorithm and implementation}
\label{sec:desc}

A pool based evolutionary algorithm can be described as an island
model \cite{whitley1997island} without topology; in fact, it is closer to the
{\em island} metaphor since migrants are sent to the {\em sea} (pool),
and come also from it, that is, the evolutionary algorithm is a
canonical one 
with binary codification, except for two steps within the cycle that
(conditionally) emit or receive immigrants. A minimum number of
evaluations for the whole algorithm is set from the beginning; we will see later on how to control when this minimum number of evaluations is reached. 

During the evolutionary loop, new individuals are selected using
3-tournament selection and generated using bit-flip mutation and
uniform crossover. Migration is introduced in the algorithm as
follows: after the population has been 
evaluated, migration might take place if the number of generations
selected to do it is reached. The best individual is sent to the pool,
and the best individual in the pool (chosen among those emitted by the
other nodes) is incorporated into the
population; if there has been no change in the best individual since
the last migration, a random individual is added to the pool, which
adds diversity to the population even if the individual fitness is not
the highest. For the time being this has no influence on the result,
but will have it later on when algorithmic tests are run. 

% La elecci\'on de un individuo aleatorio introduce diversidad en la
% poblaci\'on aunque el fitness de ese individuo sea muy malo.  
% Añadido y explicado - JJ

Migrants, if any, are incorporated into
the population substituting the worst individual, along with the offspring of the previous generation
using generational replacement with a 1-elite.  Population was set to
1000 individuals for all problems, and the minimum number of
evaluations has been four million. Several migration rates were tested
to assess its impact on performance. Besides, we introduced a 1-second
delay after migration so that workload was reduced and the Dropbox
daemon had enough time to propagate files to the rest of the
computers. This delay makes 1-computer experiments faster when less
migration is made, and will probably have to be fine-tuned in the
future. 
The results are updated at the end of the loop to check if the algorithm has
finished, that is, found the (single) solution to the problem. 

One of the advantages of this topology-less arrangement is the
independence from the number of computers participating in the
experiment, and also the lack of need from a {\em central} server,
although it can be arranged so that one of the computers starts first,
and the others start running when some file is present. Adding a new
computer, then, does not imply to arrange new connections to the
current set of computers; the only thing that needs to be done is to locate
the directory with the migrated individuals 
that is shared. 

Two representative functions have been chosen to perform the
tests; the main idea is that they took a long enough time to make
sense in a distributed environment, but at the same tame a short
enough time that experiments did not take a long time. One of them is
{\sf P-Peaks},  a multimodal problem generator proposed by 
De Jong in  \cite{dejong97using}; a {\sf P-Peaks} instance is
created by generating $P$ random $N$-bit strings where the fitness
value of a string $\vec x$ is the number of bits that $\vec x$ has
in common with the nearest peak divided by $N$.
\begin{equation}\label{eq:ppeaks}
f_{P-PEAKS}(\vec x)= \frac{1}{N} \max_{1 \leq i \leq p}\{ N -
H(\vec x,Peak_i)\}
\end{equation}
where $H(\vec x, \vec y)$ is the Hamming distance between binary
strings $\vec x$ and $\vec y$. In the experiments made in this
paper we will consider $P=100$ and $N=64$. Note that the
optimum fitness is 1.0. %Comprobar estos valores

The second function is {\sf MMDP} \cite{goldberg92massive}, which is a deceptive problem composed of 
$k$ subproblems of 6 bits each one ($s_i$). Depending of the number of
ones (unitation) $s_i$ takes the values depicted next:\\ 
\begin{center}
\begin{tabular}{ll}
$fitness_{s_i}(0) = 1.0$ & $fitness_{s_i}(1) = 0.0$ \\
$fitness_{s_i}(2) = 0.360384$ & $fitness_{s_i}(3) = 0.640576$\\
$fitness_{s_i}(4) = 0.360384$ & $fitness_{s_i}(5) = 0.0$\\
$fitness_{s_i}(6) = 1.0$   &\\
\end{tabular}
\end{center}
%%%%%%%%%%%%%%%%%%%%%%%%%%%%%%%%%%%
\begin{figure}[!htpb]
\centerline{\scalebox{0.4}{\includegraphics{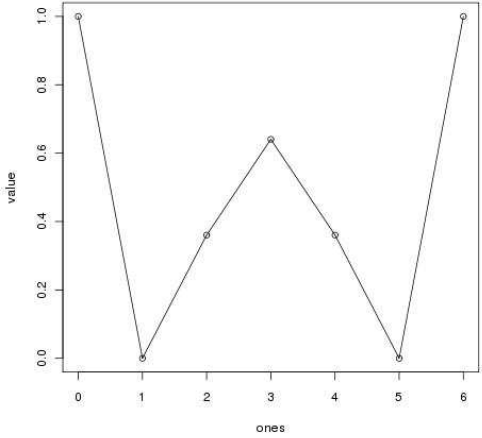}}}
\caption{Graph of the values for a single
  variable in the MMDP problem (bottom) }
\label{fgraf_mmdp}
\end{figure}
%%%%%%%%%%%%%%%%%%%%%%%%%%%%%%%%%%%

The fitness value is defined as the sum of the $s_i$ subproblems
with an optimum of $k$ (equation \ref{eq:mmdp}). 
Figure \ref{fgraf_mmdp} represents one of the variables of the function.
The number of local optima is quite large ($22^k$), while there are only
$2^k$ global solutions\footnote{The local optima occur when there are
  3 ones; off all the 64 possible combinations of six zeros and ones,
  there are 22 with exactly three ones}. 
\begin{equation}\label{eq:mmdp}
f_{MMDP}(\vec s)= \sum_{i=1}^{k} fitness_{s_i}
\end{equation}

In this paper we have used $k=20$, in order to make it difficult
enough to need a parallel solution. 

\section{Experiments and results}
\label{sec:res}

In this occasion the experiments were done in several different
computers connected also in different ways; however, computers were
added to the experiment in the same order; the problems were solved
first in a single computer, then on two and finally with four
computers. Total time, as well as the number of evaluations, were
measured. Since the end of the experiment is propagated also via
Dropbox, the number of evaluations is not exactly the one reached when
the solution is found. This number also increases with the number of
nodes. 

The only 
parameter that was changed during experiments was migration
rate. We were interested in doing this, since network performance 
will be impacted negatively with migration rate: bandwith usage (and
maybe latency) increases with the
inverse of the migration rate. On the other hand, evolutionary
performance will increase in the opposite direction: the bigger the
migration rate, the more similar to a panmictic population will be,
which might make finding the solution easier; on the other hand, it
will also decrease diversity, making the relationship between
migration rate and evolutionary/runtime performance quite complex and
worth studying. 

To keep (the rest of) the conditions uniform for one and two machines,
all parameters were fixed but for the population, which was
distributed among the machines in equal proportions: all computers
maintained a population of 1000, so that initial diversity was roughly
the same. Further experiments will have to be made keeping population
constant, but this is left for further study. 

Finally, Dropbox itself was used to check for termination conditions:
a file was written on the folder indicating the experiment had
finished; when the other node read that file, it finished too; all
nodes were kept running until the solution was found or until a
maximum number of generations were reached. That is why, in some
cases, solution is not found; the number of generations was computed
so that it was possible in a high number of cases to find out the
solution. 

The computers used in this experiment were laptops connected via the
University of Granada WiFi, they were different models, and were
running different operating systems and versions of them. The most
powerful computer was the first one; then \#2 was the second-best, and
finally numbers 3 and 4 were the least powerful ones. Since computers
run independently without synchronization checkpoints, load balancing
is automatic, with more powerful computers contributing more
evaluations to the mix, and less powerful ones contributing fewer. 

The first thing that was checked with the two problems examined
(P-Peaks and MMDP) was whether adding more computers affected the
solution rate. For P-Peaks there was no difference, independently of
migration rate and number of computers, all experiments found the
solution. However, there was a difference for MMDP, shown in Table
\ref{tab:mmdp:sols}.
\begin{table}[t]
\caption{Success rate for the MMDP problem with different number of
  nodes and migration rates. \label{tab:mmdp:sols}}
  \centering
\begin{tabular}{|l|c|c|}
\hline
\emph{Nodes} & \emph{Generations} & \emph{Success rate}\\
& \emph{between migration} & \\
\hline
1&\multirow{3}{*}{100} &0.83 \\
2&&0.95 \\
4&&1 \\
\hline 
1&\multirow{3}{*}{200}&0.70 \\
2&&0.88 \\
4&&1 \\
\hline 
1&\multirow{3}{*}{400}&0.80 \\
2&&0.90 \\
4&&1 \\
\hline
\end{tabular}
\end{table}

\begin{figure*}[ht]
\centerline{
\includegraphics[scale=0.7]{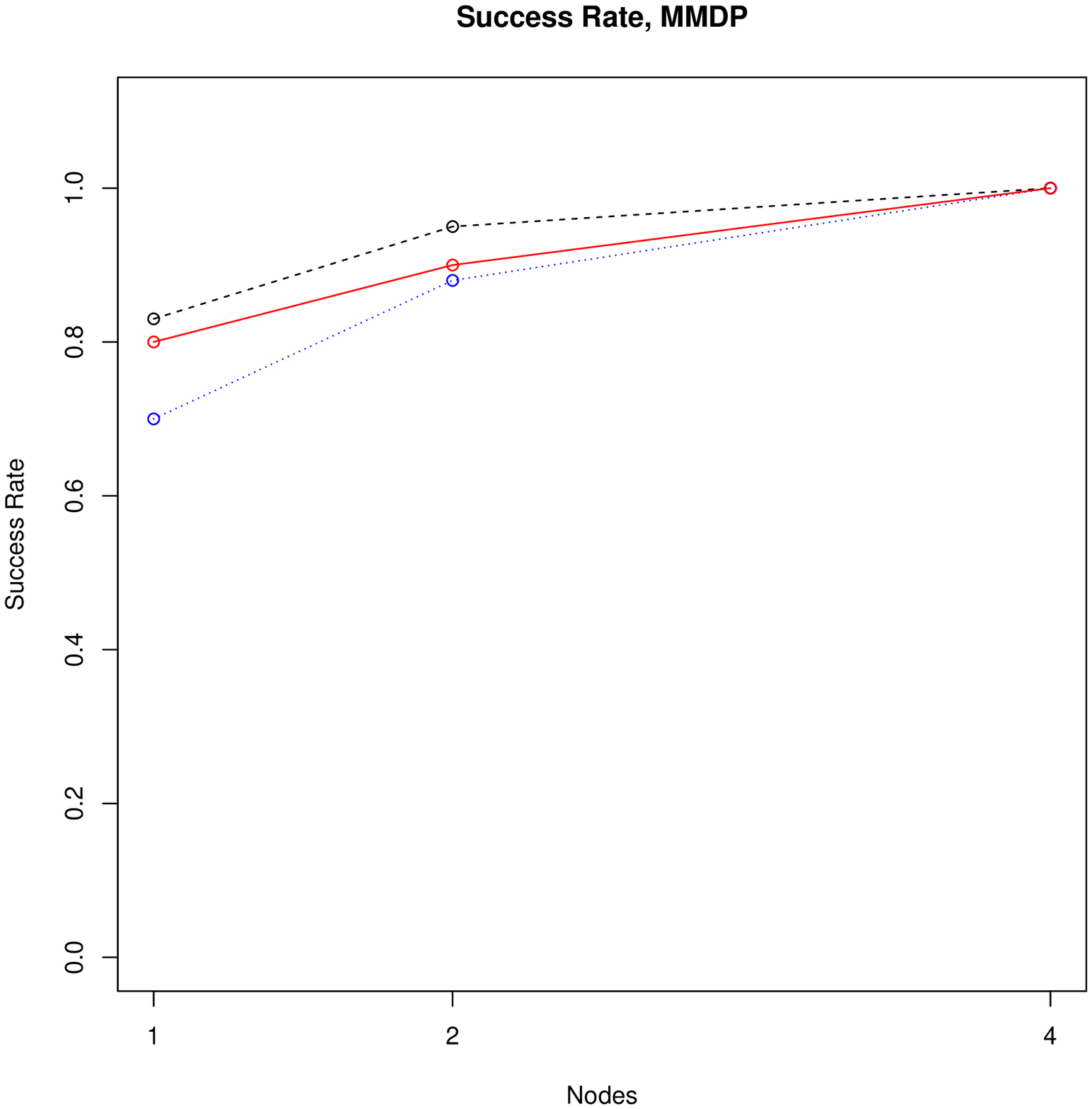}
}
\caption{Variation of the success rate in the MMDP problem with the
  number of nodes ($x$ axis) and migration rate: red-solid corresponds
  to migration rate = 400; blue-dotted to 200, and black-dashed to
  100. 
\label{fig:mmdp:sols}}
\end{figure*}
The evolution with the migration rate can also be observed in figure
\ref{fig:mmdp:sols}; as was advanced in the introduction, the
relationship is quite complex and decrease or increase do not lead to
a monotonic change of the success rate. In fact, the best success rate
corresponds to the highest migration rate (migration after 100
generations), but the second best corresponds to the lowest one
(migration after 400), which is almost akin to no migration, since
taking into account that generations run asynchronously, this might
mean that in fact on migrant from other nodes is incorporated into the
population. This result is in accordance with the {\em intermediate
  disturbance hypothesis}, proved by us previously
\cite{jj:2008:PPSN}. However, it is not clear in this case that
migration in 100 generations can be actually considered intermediate
and in 200 too high, so more experiments will have to be performed to
ascertain the optimum migration rate. 

\begin{figure*}[htbp]
\centerline{
\includegraphics[scale=0.7]{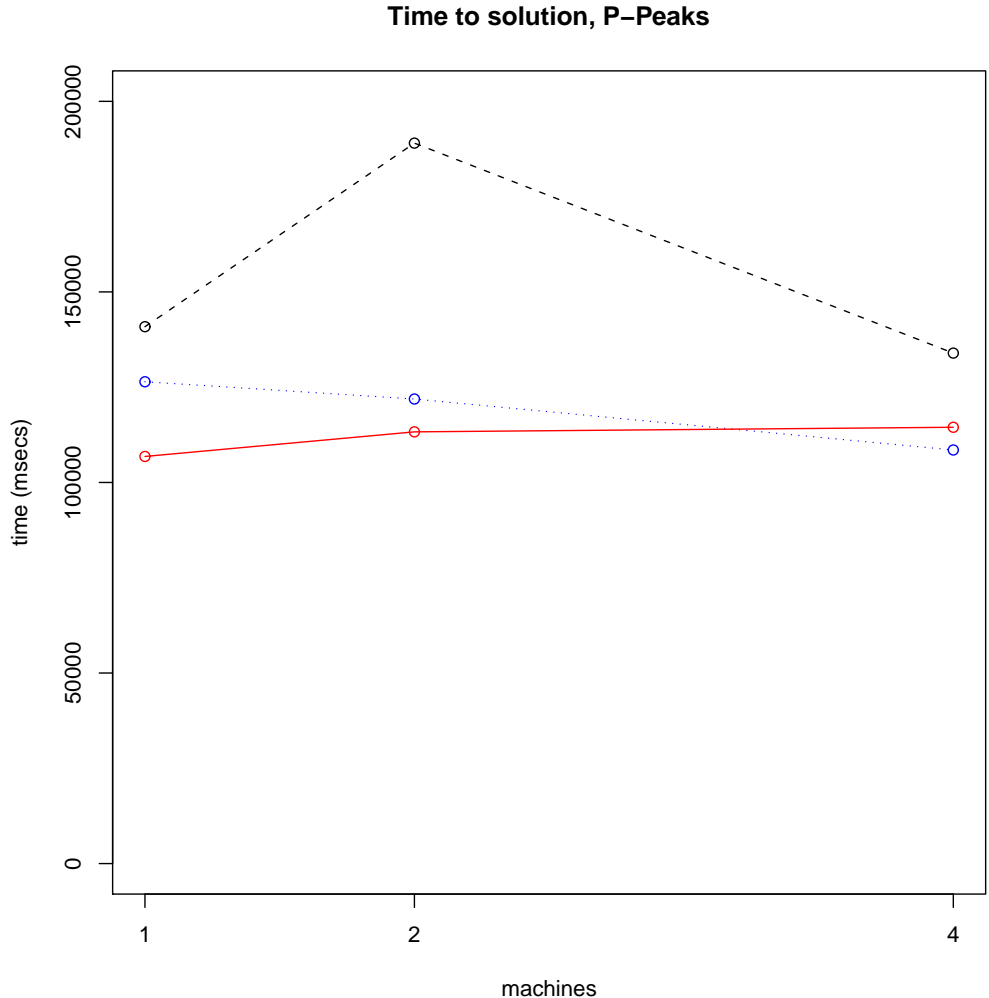}
}
\caption{Variation of the time needed to find the solution in the
  P-Peaks problem with the 
  number of nodes ($x$ axis) and migration rate: red-solid corresponds
  to migration rate = 60; blue-dotted to 40, and black-dashed to
  20. 
\label{fig:time:ppeaks}}
\end{figure*}

\begin{figure*}[htbp]
\centerline{
\includegraphics[scale=0.7]{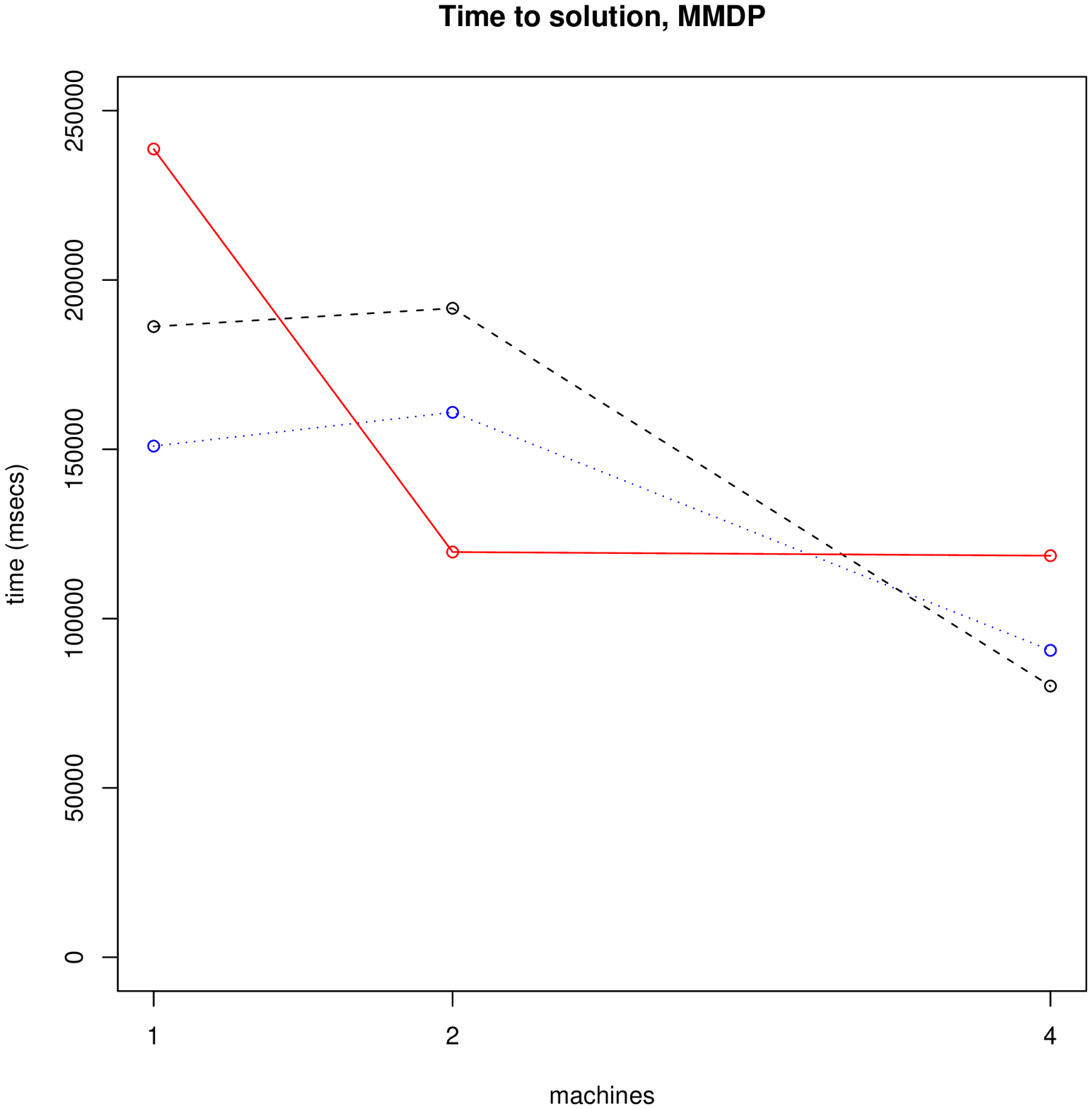}
}
\caption{Variation of the time needed to find the solution in MMDP with the 
  number of nodes ($x$ axis) and migration rate: red-solid corresponds
  to migration rate = 400; blue-dotted to 200, and black-dashed to
  100. 
\label{fig:time:mmdp}}
\end{figure*}

Thus, having proved that success rate increases with the number of
nodes, we will have to study how performance varies with it. Does the
algorithm really finds the solution faster when more nodes are added?
We have computed time only for the experiments that actually found the
solution, and plotted the results in figures \ref{fig:time:ppeaks} and
\ref{fig:time:mmdp}.

As seen above in the case of MMDP, there is not a straightforward
relationship between the migration rate and the time to solution; in
this case, the relationship between the number of computers and time
solution is also complex. If we look first at the P-Peaks experiment
in \ref{fig:time:ppeaks} we see that we obtain little time improvement
when adding more nodes to the mix. Since success rate is already 100\%
with a single computer, and the solution takes around two minutes, the
delay imposed by Dropbox implies that it is not very likely that the
migrated solutions are transmitted to the other nodes. In this case it
is the intermediate migration rate (every 40 generations) the only one
that obtains a steady decrease of time to solution. The best time is
obtained for a single node and a migration gap of 60; in general, the best
times are for the highest migration gap since the total delay induced
by migration is also the least. These results probably imply that
there must be a certain degree of complexity in the problems to take
advantage of the features in this environment. For relatively {\em
  easy} problems, which need few generations, there is nothing to
gain.

The situation varies substantially for the MMDP, as seen in figure
\ref{fig:time:mmdp}. In this case, the best result is obtained for
four nodes and the smallest mutation gap (every 100 generations,
dashed black line). However, it is interesting to observe that trend
change for two nodes in all cases, either the solution takes more time
than for a single node or it tales less than for four nodes; the
conclusion is, anyways, that the increased number of simultaneous
evaluations brought by the number of nodes eventually makes solution
faster. However, a fine-tuning of the migration gap is needed in order
to take full advantage of the parallel evaluation in the Dropbox-based
system.

\section{Conclusions and future work}
\label{sec:conc}

In general, and for complex problems like the MMDP, a Dropbox-based
system can be configured to take advantage of the paralellization of
the evolutionary algorithm and obtain reliably (in a 100\% of the
cases) solutions in less time than a single computer would. Besides,
it has been proved that it does not matter whether the new computers
added to the set are more or less powerful than the first one. In
general, however, adding more computers to a set synchronized via
Dropbox has more influence in the success rate than in the time needed
to find the solution, which seems roughly linked to the population
size, although this hypothesis will have to be tested
experimentally. On the other hand, using relatively simple problems
like P-Peaks yields no sensible improvement, due to the delay in
migration imposed by Dropbox, which implies that this kind of
technique would be better left only for problems that are at the same
time difficult from the evolutionary point of view and also slow to
evaluate. 

However, several issues remain to be studied. First, more accurate
performance measures must be taken to measure how the time needed to
find the solution in all occasions scales when new
machines are added. We will have to investigate how parameter
settings such as population size and migration gap (time passed
between two migrations) influence these measures. This paper proves
that this influence is important, but it is not clear what is the
influence on the final result. It would be also interesting to test
different migration policies affect final result, as done in
\cite{Araujo2010}, where it was found out that migrating the best one
might not be the best policy. 

An important issue too is how to interact with Dropbox so that
information is distributed optimally and with a minimal latency. In
this case we had to stop each node for a certain time (which was
heuristically found to be 1 second) to leave time for the Dropbox
daemon to distribute files. In an experiment that lasts for less than
two minutes, this can take up 25\% of the total time (per node),
resulting in an obvious drag in performance that can take many
additional nodes to compensate. A deeper examination of the Dropbox
API and a fine-tuning of these parameters will be done in order to fix
that. 

Finally, this framework opens many new possibilities for distributed
evolutionary computation: meta-evolutionary computation, artificial
life simulations, and big-scale simulation using hundreds or even
thousands of clients. The type of problems suitable for this, as well
as the design and implementation issues, will have to be
explored. Other cloud storage solutions, preferably including open
source implementations, will be also tested. Since they have different
models (synchronization daemon or user-mounted filesystems, mainly)
latency and other features will be completely different, so we expect
that performance will be affected by this. 

\section*{Acknowledgements}

% El proyecto Genil de Pedro tiene que ir exactamente as\'i, as\'i que no lo modifiqu\'eis please.
This work has been supported in part by the CEI BioTIC GENIL (CEB09-0010) MICINN CEI Program (PYR-2010-13) project and
the Andalusian Regional Government P08-TIC-03903 and P08-TIC-03928 projects.

\bibliographystyle{IEEEtran}      % Springer Computer Science
\bibliography{dropbox,geneura,ror-js,fluid,pool}

\end{document}